# High Risk of Political Bias in Black Box Emotion Inference Models

**Hubert Plisiecki***

hplisiecki@gmail.com
Institute of Psychology, Polish Academy of Sciences

ORCID: 0000-0002-5273-1716

**Paweł Lenartowicz**

pawellenartowicz@europe.com
Stowarzyszenie Otwartej Nauki

ORCID: 0000-0002-6906-7217

**Maria Flakus**

flakus.marysia@gmail.com
Institute of Philosophy and Sociology, Polish Academy of Sciences

ORCID: 0000-0002-6667-8020

**Artur Pokropek**

artur.pokropek@gmail.com
Institute of Philosophy and Sociology, Polish Academy of Sciences

ORCID: 0000-0002-5899-2917

**Abstract**

This paper investigates the presence of political bias in emotion inference models used for sentiment analysis (SA). Machine learning models often reflect biases in their training data, impacting the validity of their outcomes. While previous research has highlighted gender and race biases, our study focuses on political bias—an underexplored, pervasive issue that can skew the interpretation of text data across many studies.

We audit a Polish sentiment analysis model developed in our lab for bias. By analyzing valence predictions for names and sentences involving Polish politicians, we uncovered systematic differences influenced by political affiliations. Our findings indicate that annotations by human raters propagate political biases into the model's predictions. To mitigate this, we pruned the training dataset of texts mentioning these politicians and observed a reduction in bias, though not its complete elimination.

Given the significant implications of political bias in SA, our study emphasizes caution in employing these models for social science research. We recommend a critical examination of SA results and propose using lexicon-based systems as an ideologically neutral alternative. This paper underscores the necessity for ongoing scrutiny and methodological adjustments to ensure the reliability of the use of machine learning in academic and applied contexts.

*Keywords:* Political Bias, Emotion Inference Models, Sentiment Analysis, Social Science Research, Annotator Bias

**High Risk of Political Bias in Black Box Emotion Inference Models**

*"The bias I'm most nervous about is the bias of the human feedback raters."*

~ Sam Altman, OpenAI CEO

It is a well-documented fact that machine learning models are prone to being biased by their training data. Studies have repeatedly shown the presence of biases against various social groups, including gender and race biases, in machine learning-based sentiment analysis (SA) systems—systems that predict the positivity of text snippets [1,2,3]. These types of biases are significant from a social justice perspective, as they can exacerbate the reporting of spurious differences between social groups and affect the interpretation and outcomes of studies across various domains.

In this paper, we highlight another critical dimension of bias in SA systems: political bias, or the propagation of the political orientation of the annotators through the annotated data to the predictions of the SA model. Political bias has the potential to skew the interpretation of data across a wide range of studies, affecting societal perceptions and policymaking at a systemic level. Given that nearly every text contains some level of political nuance [4], this kind of bias can potentially influence many studies that employ SA, especially in the social sciences.

The aim of this research is to show that political bias in SA systems is substantial and pervasive. This bias not only intersects with other biases reported so far, such as gender and race biases, but also extends beyond them, rendering many research conclusions less reliable. By addressing political bias, we seek to contribute to a more comprehensive understanding of biases in SA systems and to encourage the development of mitigation strategies that enhance the reliability and fairness of SA applications.

**Emotion and Sentiment Analysis in Social Sciences**

In recent years, social scientists have increasingly recognized the profound influence of emotions across a broad range of disciplines. This interdisciplinary approach has illuminated the significant role emotions play in shaping human behavior and societal dynamics in fields such as political science [5], sociology [6,7], economics [8], anthropology [9], and organizational research [10], among others. The proliferation of text data sources—including social media, computer-based survey responses, political speeches, newspapers, online forums, customer reviews, blogs, and e-books—has provided unprecedented opportunities to examine emotions outside traditional psychological laboratory settings. Consequently, various tools have been developed to detect emotions [11,12]. As a result, research in this area has expanded rapidly.

To provide specific examples, in previous research SA been employed to predict election results [13], gauge public sentiment toward pressing social issues [14] and compare the emotional content between news sources from different ends of the political spectrum [15]. During the COVID-19 pandemic, numerous studies analyzed public sentiment based on online data [16, 17] leading to conclusions such as describing the crisis communication styles on Twitter of different Indian political leaders [18]. Similar research examined the emotional tone of the Austrian 2016 presidential election candidates [19]. From a psychological perspective, SA and emotion prediction have been used to assess suicide risk [20], automate feedback in online cognitive behavioral therapy [21], predict the subjective well-being of social media users [22], and analyze the subjective well-being of people over the past centuries [23]. All of these studies relied on sentiment analysis written text to reach scientific conclusions, showcasing the importance of this technique in current social research. However, the exact implementation of SA can vary from study to study.

Overall, there exist three main categories of sentiment analysis (SA) systems: A) dictionary-based approaches, B) large language model (LLM) approaches, and C) classical supervised predictive model approaches, referred to from now on as predictive model approaches for brevity. Dictionary-based approaches (A), also known as lexicon-based methods, rely on predefined lists of words associated with specific sentiments. These dictionaries, such as the AFINN, SentiWordNet, and LIWC [24, 25, 26], assign sentiment scores to words and phrases within a text to determine its overall sentiment. This method is straightforward and interpretable, but it can be limited by the coverage and accuracy of the dictionary, as well as by the inability to capture contextual information.

In contrast, large language model (LLM) approaches (B) leverage advanced neural networks trained on vast amounts of text data. Models such as GPT-4, LLAMA, and their derivatives can capture nuanced sentiment by understanding the context and relationships between words in a sentence and predict it in a zero-shot (without any examples to guide it), or multiple shot manner (with examples). However, their performance in emotion detection specifically falls short of the state-of-the-art (SOTA) predictive model approaches (C) [27].

The predictive model approach (C) involves training machine learning-based classifiers or regressions on labeled datasets. Techniques such as support vector machines, random forests, and deep learning models are used to predict sentiment based on features extracted from the text. These approaches are currently considered the best for analyzing emotion according to robust tests of prediction accuracy on political text datasets, as well as broader domain benchmarks [27, 28]. However, this high accuracy comes at the cost of lower interpretability and, as this study will underline, a propensity for bias.

**Bias in Predictive Models**

Bias in predictive models originates from the training data, which in the case of sentiment analysis (SA), consists of annotated text datasets. These datasets are the result of the laborious work of annotators who read through provided materials and assign emotional labels. Annotators can differ on many accounts, including age, gender, socio-economic status, psychological individual differences, and political orientation. All these differences can impact the annotation process. Studies such as Milkowski and associates [29] have shown that individual differences among annotators can significantly affect emotion annotations in text. These individual differences introduce subjectivity into data assumed to be objective, leading to inconsistencies that can skew the training and evaluation of models designed to predict emotional reactions from text. Moreover, annotation bias can result from a mismatch between authors' and annotators' linguistic and social norms, as noted by Sap and colleagues [30]. This mismatch often reflects broader social and demographic differences that can manifest in critical research areas like hate speech and abuse detection. For instance, studies by Larimore and associates [31], and Waseem [32] show that the race and gender of annotators influence not only the annotation process but also the performance of NLP models, further compounding biases.

Particularly concerning is the influence of annotators' political and ideological biases. This type of bias not only includes biases against specific social groups reported in earlier studies, but its generality makes the specific extent of its influence on SA models difficult to determine, although we expect it to be significant [1, 2, 3]. Ennser-Jedenastik and Meyer [33] report that coders of political texts often incorporate their prior beliefs about political parties into their coding decisions. For example, annotators are more likely to perceive a sentence as supporting immigration if they believe it comes from a left-wing party, regardless of the actual content. Experimental studies by van der Velden [34] show that personal characteristics of annotators, like political ideology or knowledge, interfere with their judgment of political stances. It's important to note that this interference might not be fully realized by the annotator, as previous psychological studies have shown the influence of political orientation on implicit judgments [35, 36]. Here of significant importance are the findings that show that people of different political orientations differ significantly in many annotation tasks related to political science, including emotion annotation of images [37]. This means that constructing an annotation strategy that eliminates the propagation of individual bias to SA models might be problematic. This problem parallels many similar ones in algorithm creation, where the human behavior information, on which the model is trained, falls short of the aim of the engineered algorithm. In such cases, Morewedge and associates [38] recommend auditing the models under suspicion by testing them for the presence of bias directly.

**Current Study**

In this study, we conduct a bias audit of an existing Polish sentiment analysis model developed by our lab [39] to determine whether its predicted valence readings show systematic differences based on the party affiliation of a diverse group of politicians from different political parties. We predict the valence of the names of the politicians, as well as sentences in which their names are embedded to vary based on their political affiliation (the latter were included to analyze both the direct valence towards the politicians as well as take into account the usual settings in which such a model would be used, where the name of the politicians would be a part of a specific sentence.) We regress the political affiliation of the politicians onto the sentiment readings of the model to see how much variance it can explain. To pinpoint the source of the bias, we prune the training set of any mentions of the aforementioned politicians, train a second model, and repeat the analysis. To explore the hypothesis that the model's bias is linked to the political orientation of the annotators, we administer a political orientation questionnaire to our annotators pool.

## Results

**Regression Models**

The predictive model returns the valence metric as a continuous score ranging from 0 to 100. When applied to the 24 names of politicians selected for analysis, the valence scores ranged from 42.3 to 56.6, with an average (not weighted) (M) of 49.5 and a standard deviation (not weighted) (SD) of 3.17. To examine potential bias in more natural contexts, we estimated valence for names embedded in both neutral and politically charged sentences. The mean valence was higher in neutral sentences (M = 54.4) compared to raw names (M = 49.5) and lower in politically charged sentences (M = 45.7). Interestingly, the differences in valence among politicians (measured by the standard deviation of valence) were larger for neutral sentences (SD = 4.35) compared to raw names (SD = 3.17), and smaller for politically charged sentences (SD = 1.29).

Figure 1.

*Predicted valence scores of the names of the politicians.*

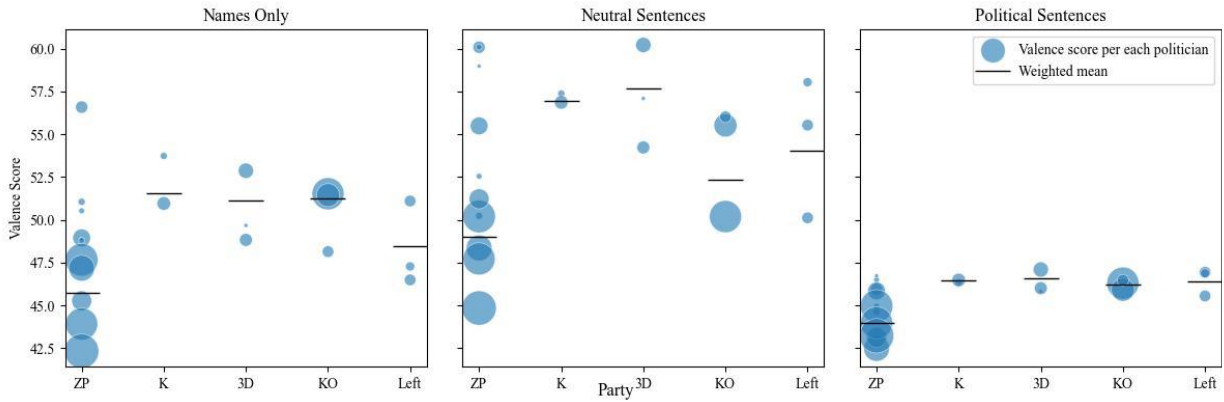

*Note*: the area of the dots corresponds to the weight (the amount of tweets containing the name of a politician)
*Abbreviations: ZP – Zjednoczona Prawica, K – Konfederacja, 3D – Trzecia Droga, KO – Koalicja Obywatelska, Left – Nowa Lewica*

Along with the visualization of the differences in predicted valence scores of politicians' names (See Figure 1.), we regressed these valence scores, as well as the predicted valence scores for the aforementioned sentences, onto the independent variables of interest (Table 1.). The fitted models with politicians' affiliation and gender (the reasoning for the inclusion of the gender confounder is driven by analyses explained in the later section Confounds) seem to describe the data well, and explain 66.5%, 52% and 66.2% of variance (R²). All of the coefficients have the same direction and similar magnitudes in all of the three models (Model 1, 2, and 3). The hypothesis of exchangeability of scores [40] could be rejected due to low p-values: $p = .008$, $p = .049$ and $p = .018$, which implies that the differences in valence are not random.

Table 1.

*Regression models – differences in valence*

|  | Dependent variable: Valence of: | | |
| --- | --- | --- | --- |
|  | Names only (1) | Neutral Sentences (2) | Political Sentences (3) |
| intercept | 45.40*** | 48.61*** | 43.89*** |
|  | (0.67) | (0.92) | (0.26) |
| 3D | 5.73** | 9.09** | 2.72** |
|  | (2.37) | (3.26) | (0.93) |
| K | 6.15* | 8.37* | 2.56* |
|  | (3.10) | (4.28) | (1.22) |
| KO | 5.83*** | 3.71* | 2.30** |
|  | (1.31) | (1.80) | (0.51) |
| Left | 3.03 | 5.46 | 2.51** |
|  | (2.56) | (3.53) | (1.01) |
| gender | 9.77** | 10.10* | 1.88 |
|  | (3.48) | (4.80) | (1.37) |
| Observations | 22 | 22 | 22 |
| R² | 0.665 | 0.520 | 0.662 |
| Adjusted R² | 0.547 | 0.370 | 0.556 |
| Residual Std. Error | 3.38 (df=16) | 5.21 (df=16) | 1.29 (df=16) |
| P-value (permutation) | 0.008 | 0.049 | 0.018 |

*Note*: *p<0.1; **p<0.05; ***p<0.01 (*t*-test)
Zjednoczona Prawica (ruling party) as intercept, gender: woman=1, man=0
*Abbreviations:* K – Konfederacja, 3D – Trzecia Droga, KO – Koalicja Obywatelska, Left – Nowa Lewica

**Confounds**

The association of political affiliation and valence was significantly stronger than between valence and the confounders. This is evidenced by comparing R² of regression on valence in raw names (Model 1, Table 2.) and political affiliation (R² = .49), and models with only confounds as independent variables. The model including only gender reared R² = .109 (statistics that relate to gender should be interpreted carefully since there are only 2 women in our sample), trust towards

Zpolitician achieved $R^2 = .195$, and the mean valence of mentions in which a given politician appeared resulted in $R^2 = .175$. (These models are available in the Appendix)

To find the model that best describes the data, we compare adjusted $R^2$ with different sets of potential confounds. Since the model with affiliation and gender as independent variables has the highest adjusted $R^2$, this set of independent variables is used in other models. (See Table 2.)

Table 2.

*Regression models – inspecting confounders*

|  | Dependent variable: Valence in raw names | | | | |
| --- | --- | --- | --- | --- | --- |
|  | (1) | (2) | (3) | (4) | (5) |
| intercept | 45.76*** | 45.40*** | 45.52*** | 45.76*** | 46.06*** |
|  | (0.78) | (0.67) | (0.69) | (1.38) | (1.43) |
| 3D | 5.36* | 5.73** | 5.55** | 5.14 | 4.67 |
|  | (2.80) | (2.37) | (2.39) | (3.14) | (3.19) |
| K | 5.79 | 6.15* | 5.86* | 4.91 | 4.01 |
|  | (3.67) | (3.10) | (3.14) | (5.23) | (5.35) |
| KO | 5.47*** | 5.83*** | 5.40*** | 5.69*** | 5.16*** |
|  | (1.54) | (1.31) | (1.41) | (1.43) | (1.55) |
| Left | 2.67 | 3.03 | 3.01 | 2.71 | 2.54 |
|  | (3.03) | (2.56) | (2.86) | (2.85) | (2.87) |
| gender |  | 9.77** | 8.72* | 9.61** | 8.40** |
|  |  | (3.48) | (3.71) | (3.63) | (3.88) |
| trust |  |  | 0.71 |  | 0.76 |
|  |  |  | (0.80) |  | (0.83) |
| mentions |  |  |  | 0.037 | 0.055 |
|  |  |  |  | (0.124) | (0.127) |
| Observations | 22 | 22 | 22 | 22 | 22 |
| $R^2$ | 0.485 | 0.655 | 0.672 | 0.657 | 0.676 |
| Adjusted $R^2$ | 0.364 | 0.547 | 0.541 | 0.520 | 0.514 |
| Residual Std. Error | 3.88 (df=17) | 3.38 (df=16) | 3.39 (df=15) | 3.35 (df=15) | 3.31 (df=14) |
| P-value (permutation) | 0.051 | 0.008 | 0.021 | 0.017 | 0.039 |

*Note*: *$p<0.1$; **$p<0.05$; ***$p<0.01$ (*t*-test)
Zjednoczona Prawica (ruling party) as intercept, gender: woman=1, man=0, trust: normalized trust scores, mentions: mean valence of annotated text, in which politician was mentions in 0-100 scale
*Abbreviations:* K – Konfederacja, 3D – Trzecia Droga, KO – Koalicja Obywatelska, Left – Nowa Lewica

**The Modified Model**

In the model modified (See Table 3.) by pruning texts containing mentions of our set of politicians, the relationships between affiliation and valence decreased significantly, but bias was still present in the model with raw names (Model 1., Table 3.) It should be noted that not all mentions affecting the model could be pruned, for example, the most mentioned politician is Jarosław Kaczyński, but in the dataset there are tweets mentioning his twin brother, Lech Kaczyński, the former president and member of the same party.

Table 3.

*Regression models – modified model (text pruning)*

|  | *Dependent variable: Valence of: (modified model)* | | |
|---|---|---|---|
|  | raw names (1) | neutral sentences (2) | political sentences (3) |
| intercept | 49.48*** | 53.85*** | 45.09*** |
|  | (0.51) | (1.14) | (0.34) |
| 3D | 3.70* | 5.93 | 0.77 |
|  | (1.79) | (4.00) | (1.20) |
| K | 2.69 | 4.64 | 0.41 |
|  | (2.35) | (5.26) | (1.58) |
| KO | 1.55 | 0.87 | 0.75 |
|  | (0.99) | (2.21) | (0.66) |
| Left | 0.76 | 3.38 | 0.90 |
|  | (1.94) | (4.34) | (1.30) |
| gender | 6.47** | 7.74 | 0.77 |
|  | (2.63) | (5.90) | (1.77) |
| Observations | 22 | 22 | 22 |
| R² | 0.421 | 0.224 | 0.104 |
| Adjusted R² | 0.240 | -0.019 | -0.176 |
| Residual Std. Error | 3.02 (df=16) | 4.87 (df=16) | 1.10 (df=16) |
| P-value (permutation) | 0.076 | 0.101 | 0.202 |

*Note*: *p<0.1; **p<0.05; ***p<0.01 (*t*-test)
Zjednoczona Prawica (ruling party) as intercept, gender: woman=1, man=0,
*Abbreviations:* K – Konfederacja, 3D – Trzecia Droga, KO – Koalicja Obywatelska, Left – Nowa Lewica

**Questionnaire Results**

For the "The government should take action to reduce income differences" item, 4 people (26.7%) responded that they "Definitely agree"; "Agree" - 1 (6.7%); "Neither agree nor disagree" - 5 (33.3%); "Disagree" - 5 (33.3%). Nobody chose the option "Definitely disagree". For the second item, "People with a homosexual orientation, gays and lesbians, should have the freedom to arrange their lives according to their own beliefs" 13 people responded with "Definitely agree" (86.7%), and 2 people that they "Agree" (13.3%), these were the only two options picked by the annotators.

## Discussion

In the current study we have shown that a supervised model trained on annotations created by expert annotators in their domain shows signs of political bias with regards to well-known politicians. The impact of this bias depends on the analytical context. While using our original model, introducing a politically charged surname can alter the sentiment score of single text snippet by up to 6 points on a 1-100 scale or by 0.5 in terms of Cohen's d. Given the effect size, datasets with minimal politically charged content may experience only minor bias issues. However, when comparing political groups—as in regression analyses across different political parties—this bias becomes systematic. For instance, the difference between groups, when compared to the pooled standard deviation, is 5.47 versus 3.53 (Cohen's d = 1.55) for raw names, or 2.30 versus 1.23 (Cohen's d = 1.87) for modified names. This suggests that bias could be even more pervasive in datasets with greater political content and in inter-group analyses.

Interestingly, while the variance in sentiment predictions of raw names, and neutral sentences was higher than in the case of political sentences, which most probably relates to the higher level of emotional signal present in the latter as the words in political sentences contained more emotionally charged language such as "support", or "against", the systematicity of the bias present in political sentences counteracted this effect as the regression models conducted on each of these conditions explained similar levels of variance (0.665, 0.52, and 0.662 consecutively). This may suggest that the presence of additional political context in the predicted text could prime the model to focus on the political information, thereby making the bias more systematic.

Of note is the fact that this bias was not explained by the publics' trust towards politicians, indicating that it does not reflect the general society's political preferences, but rather those of a selected nonrepresentative group. Similarly, the mean valence of the training data tweets that included the names of the politicians did not explain this bias either. This rules out the hypothesis that the bias comes from a systematic difference in valence between how the politicians were portrayed in text. Another argument against the influence of the linguistic bias is the moderate intraclass correlation coefficient (0.6) indicating limited agreement between the annotators which further undermines the possibility of the bias being inherent to the text of the training dataset, and not to the subjective perception of the text by the annotators.

The modified model, trained on a dataset pruned of texts containing politicians' names, exhibited significantly lower bias than the primary model. This suggests that at least a substantial part of the bias can be attributed to the annotations made by the annotation team. This, however, does not indicate that pruning the names of the politicians eradicates all kinds of biases that political orientation might result in. Additionally, the model cannot be fully isolated from the influence of certain mentions that may affect its output. For instance, while identifying mentions of Jarosław Kaczyński, snippets related to his twin brother, Lech Kaczyński, might be included, potentially influencing the model's predictions. More indirect sources of bias might also be present. Moreover, certain word associations may be embedded in the model's initial architecture before training for emotion detection, and this pre-existing knowledge could interact with the annotations producing harder to eradicate bias. Given these limitations, along with broader challenges in practical application, we do not recommend pruning as a method for bias mitigation. Furthermore, the instructions given to the annotators, which prompted them to estimate the "positivity/negativity that they read in each text" rather than their emotional reactions to it, bias still propagated into the annotated dataset in an implicit manner. Instances of such implicit propagation of political orientation have been documented in previous psychological research [35, 36].

The most likely explanation for this effect is that when annotators encountered a text mentioning a politician, they tended to label it with a positive or negative bias aligned with their own political orientation. However, during training, this bias was treated as ground truth. As a result, any difference in valence between this text and a similar one without the politician's name was attributed by the model to the politician's name itself. Repeated exposure to this pattern led to a systematic bias. By removing texts containing such biased annotations, we therefore reduced this bias.

Considering the political questionnaire results, there is some evidence indicating that the political beliefs of the annotators may correlate with the model's bias, though this correlation is not definitive. We know that annotators have varying opinions on economic issues and tend to be progressive on social issues. If these factors were the main sources of bias, the bias towards a progressive party (Nowa Lewica) should differ significantly from that towards the most conservative party (Konfederacja). However, this is not the case as the results consistently show a lower score for the ruling party (Zjednoczona Prawica) and a higher valence for all opposition parties, including Konfederacja. This could indicate an anti-government sentiment stronger than political leanings.

Alternatively, existing evidence that people's political attitudes may not fully align with their supported parties' programs might prompt a different interpretation. Polish voter survey results [41] reveal that opposition party voters' views are more similar to each other than to those of their party leaders and programs. This could explain the relative positivity of the Konfederacja members, despite the strong progressive leaning of the political questionnaire results. Furthermore, Zjednoczona Prawica's voters are shown to be more culturally conservative and economically leftist compared to voters of other parties and the implications of the party's program. The opposite goes for the questionnaire respondents, being strongly progressive, and inconsistent when it comes to economic issues. While the nature of the collected data does not permit drawing specific conclusions as to how exactly the political leanings of the annotators affect annotation results, previous studies on the manner support the general hypothesis of the propagation of the political orientation of annotators to the annotation results [35, 36, 37].

The existence of political bias in the model has thus been clearly documented, and so has its causal connection to the training data. This means not only that the annotations made by humans can lead to biased models, but also raises the very real possibility that their bias might have spread to more concepts in the dataset. If people implicitly propagate their political orientation towards social groups [1, 2, 3] as well as specific politicians as proven by the current study, the only thing standing in the way of abstract concepts being affected by the same type of bias is the ability of the model to pick up on it.

As language models become more advanced, their understanding of language becomes gradually less reliant on specific entities which they pick up from the text as in the case of for example Naive Bayes algorithms [42], and more reliant on relations between abstract concepts. This is evidenced by the distributed nature of the information that large language models and other transformer-reliant architectures use, through the mechanisms of attention, to generate their outputs [43], as well as by the recent LLM interpretability research showcasing the crystallization of abstract concepts within the inner layers of these models [44]. This means that as models improve, the propensity of the models being biased towards specific abstract concepts such as for example anarchism, or democracy might increase, given that such bias will be present in the training data, which is likely. Furthermore, the inspection of these kinds of distributed, conceptual biases will require new, more complex methods of bias detection.

Given the biases that have been already uncovered in SA models, as well as those more abstract that can lurk in the shadows, yet unidentified, we discourage the use of such models for research and advise caution in interpreting the results of those that have already used them. To stress the kinds of problems their use can lead to, let's go back to the examples of research performed with the use of SA systems. The analysis of the sentiment towards social issues might be biased towards the sentiment of the annotators team [14]. Similarly, when comparing emotional content of news sources, the same propagation of bias can occur [15], directly biasing the conclusions. This problem of propagation of bias directly biases studies that apply their SA systems to compare different groups of texts in terms of emotionality. When trying to predict something using SA scores, like in the case of predicting election results, assessing suicide risk, or subjective wellbeing, the effectiveness of the predictive model can be influenced by the beliefs of the annotator group, leading to replication issues [22, 20, 13]. At the same time, when creating customer facing solutions such as automating feedback in online cognitive behavioral therapy one has to consider that annotator biases might lead to people with different political predispositions receiving different standards of care, however here the influence is not as clear cut as in earlier cases.

However, not all of the studies mentioned in this paper are affected by this bias, as many of them have relied on lexicon-based SA systems, forgoing the increased accuracy of the predictive models. As these approaches depend on lists of emotionally loaded words which are not ideologically relevant, annotated separately, and without any contexts, they are significantly less susceptible to propagating the bias of their annotators. Furthermore, any bias that they do propagate can be clearly read from the word annotations themselves, therefore researchers that want to buttress their analysis against specific biases can directly check for them within the lexicon and correct them there and then. The same task is orders of magnitude more complicated when using black box ML models and becomes even more complicated when the bias concerns concepts rather than entities.

The higher accuracy of transformer-based, and other predictive models could be therefore traded in for the less accurate, but more bias- safe lexicon-based systems. However, given that the drop in performance when using lexicon approaches is quite severe this might not be a preferred solution for some researchers [28]. Future research could also focus on creating emotion prediction ML models that do not propagate training bias. In the case where authors choose to use ML based SA systems anyway, we recommend them to take the possibility of different types of potential biases into consideration when analyzing their results, and if possible, to corroborate their results using a lexicon-based system.

The alternative in the form of picking the annotation team so that it is balanced with regards to all of the individual differences such as political orientation and others that could influence their annotations is problematic as 1. it is as of yet not clear which differences could play a role in the annotation process 2. balancing a large number of them would require a very large annotation team which would be very resource intensive. Nonetheless in very specific applications where the nature of the bias relevant to a given experiment can be directly pinpointed such solutions might be viable.

In conclusion, the current paper shows that supervised models trained on datasets annotated by humans are susceptible to showing the same biases as annotators, despite the annotation instructions being phrased in a way that should avoid the propagation of such bias. This result should be taken into consideration when conducting and interpreting sentiment analysis research in the political science sphere and beyond. We therefore recommend the research community to perceive machine learning based sentiment analysis models as biased until proven otherwise and consider exploring alternative approaches.

The main limitation of the current study is its focus on a single sentiment analysis model and a specific dataset largely composed of political texts in Polish. While these conditions are ideal for exploring political bias within the context of Polish politics, the generalizability of the findings cannot be stated with certainty, although should be taken into consideration. We recommend the researchers that are in doubt of whether our results extend to their models to replicate our findings before using them. Additionally, the sample size of politicians and the specific sentences used to assess bias were relatively small, which may limit the robustness of our regression analyses. The political orientation questionnaire administered to annotators also had a limited scope, potentially overlooking other ideological dimensions that could influence annotation behavior. Future research should aim to replicate these findings across diverse datasets, expand the number of annotators and the range of their political orientations, and explore the interaction between different types of bias in sentiment analysis models. This would not only help validate the current results but also enhance our understanding of how various biases can coexist and interact within AI models used in social science research.

## Methods

**The Prediction Model**

*Model Training Data*

The model has been trained on a training set sampled from a comprehensive database of Polish political texts from social media profiles (i.e., YouTube, Twitter, Facebook) of 25 journalists, 25 politicians, and 19 non-governmental organizations (NGOs). The complete list of the profiles is available in the Appendix. For each profile, all available posts from each platform were scraped (going back to the beginning of 2019). In addition, we also included texts written by "typical" social media users, i.e., non-professional commentators of social affairs. Our data consists of 1246337 text snippets (Twitter: 789490 tweets; YouTube: 42252 comments; Facebook: 414595 posts).

As transformer models have certain limits, i.e., their use imposes limits on length, we implemented two types of modification within the initial dataset. First, since texts retrieved from Facebook were longer than the others, we have split them into sentences. Second, we deleted all texts that were longer than 280 characters.

The texts were further cleaned from social media artifacts, such as dates scrapped alongside the texts. Next, the *langdetect* [45] software was used to filter out text snippets that were not written in Polish. Also, all online links and usernames in the texts were replaced with "_link_" and "_user_", respectively, so that the model does not overfit the sources of information nor specific social media users.

Because most texts in the initial dataset were emotionally neutral, we filtered out the neutral texts and included only those that had higher emotional content in the final dataset. To filter the neutral snippets, the texts were stemmed and subjected to a lexicon analysis [46] using lexical norms for valence, arousal, and dominance - the three basic components of emotions. The words in each text were summed up in terms of their emotional content extracted from the lexical database and averaged to create separate metrics for the three emotional dimensions. These metrics were then summed up and used as weights to choose 8000 most emotionally loaded texts for the final training dataset. Additionally, 2000 texts were selected without weights to ensure the resulting model could process both neutral and emotional texts. The proportions of the texts coming from different social media platforms reflected the initial proportions of these texts, resulting in 496 YouTube texts, 6105 Twitter texts, and 3399 Facebook texts, totaling 10,000 texts.

*Annotators and Political Orientation Questionnaire*

The final dataset consisting of 10,000 texts was annotated by 20 expert annotators (age: $M = 23.89$, $SD = 4.10$; gender: 80% female) with regards to six emotions: happiness, sadness, disgust, fear, anger, and pride, as well as to two-dimensional emotional metrics of valence and arousal, using a 5-point Likert scale. All annotators were well-versed in Polish political discourse and were students of Psychology (70% of them were graduate students, which in the case of Polish academic education denotes people studying 4th and 5th year). Thus, they underwent at least elementary training in psychology. Each text was annotated by 5 randomly picked annotators. The inter annotator reliability as measured by the intraclass correlation coefficient (ICC(1)) for valence measured 0.60 indicating moderate reliability [47].

Since valence and arousal might not have been familiar to annotators, before the formal annotation process began, all annotators were informed about the characteristics of valence and arousal. General annotation guidelines were provided to ensure consistency and minimize subjectivity. For the purpose of annotating valence of texts, the annotators were given the following instruction:

English translation (An in-depth description of the annotation process is available in the Appendix):

*"Go back to the text you just read. Now think about the sign of emotion (positive / negative) and the arousal you read in a given text (no arousal / extreme arousal). Rate the text on these emotional dimensions."*

Five months after the annotation process took place, the annotators received a political orientation questionnaire in Polish, which consisted of two items: "The government should take action to reduce income differences", and "People with a homosexual orientation, gays and lesbians, should have the freedom to arrange their lives according to their own beliefs", to which they had to respond with whether they "Definitely agree", "Agree", "Neither agree nor disagree", "Disagree", or "Definitely disagree" with them. These questions were sourced from the European Social Survey [48] and were meant to capture as much variance in the political views of the annotators as possible, while at the same time not requiring them to complete a long and tedious questionnaire. The questionnaire was anonymous and not obligatory, as it was no longer a part of the annotation process for which they were hired. 15 responses total have been received, which constitutes 75% of the original annotation team.

*Model Training*

For model training, we have considered two alternative base models: the Trelbert transformer model developed by a team at DeepSense [49], and the Polish Roberta model [50]. The encoders of both models were each equipped with an additional regression layer with a sigmoid activation function. The models have been trained to predict each of the six emotion intensities, as well as valence, and arousal. The maximum number of epochs in each training run was set to 100. At each step, we computed the mean correlation of the predicted metrics with their actual values on the evaluation batch, and the models with the highest correlations on the evaluation batch were saved to avoid overfitting. We used the MSE criterion to compute the loss alongside the AdamW optimizer with default hyperparameter values. Both of the base models were then subjected to a Bayesian grid search using the WandB platform [51] with the following values: dropout - 0; 0.2, 0.4, 0.6; learning rate - 5e-3, 5e-4, 5e-5; weight decay - 0.1, 0.3, 0.5; warmup steps - 300, 600, 900. The model which obtained the highest correlation relied on the Roberta transformer model and had the following hyperparameters: dropout = 0.6; learning rate = 5e-5; weight_decay = 0.3. Its average accuracy on the test set is $r = 0.80$, and $r = 0.87$ valence, which is the main metric analyzed in the current study as it shows the estimated general positivity of the analyzed text.

**Bias Testing**

*Stimuli*

As stimuli for testing the bias hypothesis, to limit our arbitrary choice of stimuli, we used the names of 24 well-known Polish politicians who appeared in the November and October 2023 trust polls [52, 53, 54]. The politicians were assigned to 5 political parties/coalitions on the basis of their affiliation or because they were candidates of that party/coalition. These parties/coalitions are Zjednoczona Prawica, which is right-wing and was the ruling coalition, Trzecia Droga, Koalicja Obywatelska, Nowa Lewica, which were centre-right, centre and left opposition respectively. The fifth party was Konfederacja, which was a right to far right opposition. These coalitions cover 96.25% of the total votes in the November 2023 parliamentary elections [55].

In addition to predicting the valence of politicians' names themselves, we also embedded them in neutral sentences and sentences with political context to estimate how much their presence changes the valence predicted by the model. Both the politicians' names and the sentences are available in the appendix.

*Corpus Modification*

To identify the potential source of the model's bias, we locate the texts in the training set that contain the surnames of these politicians. We then manually review these texts to see if they are referring to a particular politician. There are 459 of these texts in total, with a range of 71 to 0 and a median of 8.5 per politician. We then prune the training set of these texts and train a second model with the same training parameters to estimate the degree to which their presence influences the model's bias. The training set contained 7999 texts before pruning, which means that the pruned texts constitute below 6% of its size.

**Statistical Analyses**

To test for the presence of bias, we examine where there are noticeable differences in the valence of politicians' names and where they can be explained by the politicians' political affiliations. For this purpose, we build several regression models. As dependent variables, we use the valence score from the original model, the same score from the modified model (trained on the pruned corpus), and the differences in valence between the final and the modified model. The models return the valence score as continuous variables ranging from 0 to 1, which we chose to then recalculate on a 0-100 scale for better readability.

As independent variables we use the politicians' affiliation, and potential confounders: their gender, trust towards them (from the same trust surveys as the names of politicians) and mean annotated valence of texts in which these politicians appear, recalculated to 0-100. We included a trust "score" as a proxy to control for the general favourability of each politician, to separate it from nonrepresentative political bias. This allows us to account for positive or negative feelings about a politician that may stem from their popularity or personal traits rather than political alignment. Additionally, the mean valence of the training dataset snippets with the politicians' names was included to test the possibility that the bias of the model stems from the politicians names cooccurring with systematically biased language. By incorporating both trust and mean annotated valence scores, we aimed to rule out alternative explanations for differences in valence beyond the annotators' political bias.

The trust surveys were decoded as 5-point Likert scales [54] or 3-point Likert scales [52, 53]. Responses "I don't know" and "difficult to say" were recoded as neutral. For each survey, a normalized score was calculated, and the mean of these normalized scores was included in the analysis. Mean annotated valence scores were derived from texts that were later pruned, see 'Corpus Modification', and recalculated to 0-100 scale.

For the regression models we use the weighted least squares method [56], weighted by the number of mentions of a given politician. Due to the weighting process, two politicians without any mentions in training data were excluded. To test the null hypothesis of lack of correlation between political affiliation and bias in the models, we conducted the permutation tests on the observed valence (Manly, 1997) for each model, with 100,000 random assignments. This method guarantees robustness and decent statistical power [57]. This method could be vulnerable to extreme outlier in dependent variables, which is not a problem in this study, due to the categorical or bounded character of dependent variables used in this study. The QQ-plots of the model residuals are included in the appendix. Due to small sample sizes for affiliations, parametric (assuming normal distributions) confidence intervals are calculated.

**Data availability statement**

The code used in the current study is available at the github repository https://github.com/hplisiecki/political-model-bias and the OSF directory https://osf.io/q8bes/. The base model used for the current study is available at https://huggingface.co/hplisiecki/polemo_intensity, while the modified model is available at the aforementioned OSF repository.

**Funding**

This research is funded by a grant from the National Science Centre (NCN) 'Research Laboratory for Digital Social Sciences' (SONATA BIS-10, No. UMO-020/38/E/HS6/00302).

**Disclosure Statement**

**Political Bias in Emotion Inference** 29systematic review. Expert Systems with Applications, 167, 114155.

https://doi.org/10.1016/j.eswa.2020.114155

17. Wang, J., Fan, Y., Palacios, J., Chai, Y., Guetta-Jeanrenaud, N., Obradovich, N., Zhou, C., & Zheng, S. (2022). Global evidence of expressed sentiment alterations during the COVID-19 pandemic. Nature Human Behaviour, 6(3), 349–358. https://doi.org/10.1038/s41562-022-01312-y

18. Kaur, M., Verma, R., & Otoo, F. N. K. (2021). Emotions in leader's crisis communication: Twitter sentiment analysis during COVID-19 outbreak. Journal of Human Behavior in the Social Environment, 31(1–4), 362–372. https://doi.org/10.1080/10911359.2020.1829239

19. Kušen, E., & Strembeck, M. (2018). Politics, sentiments, and misinformation: An analysis of the Twitter discussion on the 2016 Austrian Presidential Elections. Online Social Networks and Media, 5, 37–50. https://doi.org/10.1016/j.osnem.2017.12.002

20. Glenn, J. J., Nobles, A. L., Barnes, L. E., & Teachman, B. A. (2020). Can Text Messages Identify Suicide Risk in Real Time? A Within-Subjects Pilot Examination of Temporally Sensitive Markers of Suicide Risk. Clinical Psychological Science, 8(4), 704–722. https://doi.org/10.1177/2167702620906146

21. Provoost, S., Ruwaard, J., Van Breda, W., Riper, H., & Bosse, T. (2019). Validating automated sentiment analysis of online cognitive behavioral therapy patient texts: An exploratory study. Frontiers in Psychology, 10, 1065.

22. Chen, L., Gong, T., Kosinski, M., Stillwell, D., & Davidson, R. L. (2017). Building a profile of subjective well-being for social media users. PLOS ONE, 12(11), e0187278. https://doi.org/10.1371/journal.pone.0187278

**Political Bias in Emotion Inference**                                                                 333 Sonnet. Transformer Circuits Thread. https://transformer-circuits.pub/2024/scaling-monosemanticity/index.html

45. Danilak, M. M. (2021). langdetect: Language detection library ported from Google's language-detection. (1.0.9) [Python; OS Independent]. https://github.com/Mimino666/langdetect

46. Imbir, K. K. (2016). Affective Norms for 4900 Polish Words Reload (ANPW_R): Assessments for Valence, Arousal, Dominance, Origin, Significance, Concreteness, Imageability and, Age of Acquisition. Frontiers in Psychology, 7, 1081. https://doi.org/10.3389/fpsyg.2016.01081

47. Koo, T. K., & Li, M. Y. (2016). A Guideline of Selecting and Reporting Intraclass Correlation Coefficients for Reliability Research. Journal of Chiropractic Medicine, 15(2), 155–163. https://doi.org/10.1016/j.jcm.2016.02.012

48. European Social Survey. (2020). ESS round 10 source questionnaire. ESS ERIC Headquarters c/o City, University of London.

49. Szmyd, W., Kotyla, A., Zobniów, M., Falkiewicz, P., Bartczuk, J., & Zygadło, A. (2023). TrelBERT: A pre-trained encoder for Polish Twitter. In J. Piskorski, M. Marcińczuk, P. Nakov, M. Ogrodniczuk, S. Pollak, P. Přibáň, P. Rybak, J. Steinberger, & R. Yangarber (Eds.), Proceedings of the 9th Workshop on Slavic Natural Language Processing 2023 (SlavicNLP 2023) (pp. 17–24). Association for Computational Linguistics. https://doi.org/10.18653/v1/2023.bsnlp-1.3

50. Dadas, S. (2020). Sdadas/polish-roberta [Python]. https://github.com/sdadas/polish-roberta (Original work published 2020)

# Appendix

## 1. Annotation Process

The final dataset consisting of 10,000 texts was annotated by 20 expert annotators (age: M = 23.89, SD = 4.10; gender: 80% female). All annotators were well-versed in Polish political discourse and were students of Psychology (70% of them were graduate students, which in the case of Polish academic education denotes people studying 4th and 5th year). Thus, they underwent at least elementary training in psychology.

The entire annotation process lasted five weeks. Each week, every annotator was given five sets of texts (out of 100 sets with 100 randomly assigned sentences each) that should be annotated in the given week. The sets were randomly assigned to annotators, taking into account the general assumption that five different annotators should annotate each set. Generally, annotators simultaneously annotated no more than 500 texts each week, preventing them from cognitive depletion's negative effects.

Annotators labeled each text based on five basic emotions: happiness, sadness, anger, disgust, and fear. In addition, annotators were asked to label the texts with regard to additional emotion, namely pride, and two general dimensions of emotions: valence and arousal. In all cases, annotators used a 5-point scale (in the case of emotions: 0 = *emotion is absent*, 4 = *very high level of emotion*; in the case of valence and arousal, we used a pictographic 5-point scale provided in Supplementary materials).

Since two additional emotional dimensions might not have been familiar to annotators, before the formal annotation process began, all annotators were informed about the characteristics of valence and arousal (note that we did not provide formal definitions of basic emotions). General annotation guidelines were provided to ensure consistency and minimize subjectivity.

**2. Social Media Profiles**

The Social Media profiles scraped to generate the training dataset:

A) Journalists:

Adrian Klarenbach, Agnieszka Gozdyra, Bartosz T. Wieliński, Bartosz Węglarczyk, Bianka Mikołajewska, Cezary Krysztopa, Daniel Liszkiewicz, Dawid Wildstein, Dominika Długosz, Dominika Wielowieyska, Ewa Siedlecka, Jacek Karnowski, Jacek Kurski, Jacek Nizinkiewicz, Janusz Schwertner, Jarosław Olechowski, Konrad Piasecki, Krzysztof Ziemiec, Łukasz Bok, Łukasz Warzecha, Magdalena Ogórek, Magdalena Rigamonti, Marcin Gutowski, Marcin Wolski, Michał Karnowski, Michał Kolanko, Michał Rachoń, Miłosz Kłeczek, Paweł Żuchowski, Piotr Kraśko, Piotr Semka, Radomir Wit, Rafał Ziemkiewicz, Renata Grochal, Robert Mazurek, Samuel Pereira, Szymon Jadczak, Tomasz Lis, Tomasz Sakiewicz, Tomasz Sekielski, Tomasz Sommer, Tomasz Terlikowski, Wojciech Bojanowski, Agaton Koziński, Piotr Witwicki, Jacek Tacik, Magdalena Lucyan, Agata Adamek, Kamil Dziubka, Jarosław Kurski, Dorota Kania, Ewa Bugala, Zuzanna Dąbrowska, Karol Gac, Marcin Tulicki, Marzena Nykiel, Jacek Prusinowski, Paweł Wroński

B) Politicians:

Donald Tusk, Andrzej Duda, Rafał Trzaskowski, Mateusz Morawiecki, Sławomir Mentzen, Janusz Korwin-Mikke, Grzegorz Braun, Szymon Hołownia, Radosław Sikorski, Krzysztof Bosak, Władysław Kosiniak-Kamysz, Borys Budka, Artur E. Dziambor, Marek Belka, Leszek Miller, Mariusz Błaszczak, Roman Giertych, Franek Sterczewski, Konrad Berkowicz, Marek Jakubiak, Michał Szczerba, Przemysław Czarnek, Zbigniew Ziobro, Krzysztof Brejza, Leszek Balcerowicz, Izabela Leszczyna, Klaudia Jachira, Janusz Piechociński, Patryk Jaki, Robert Biedroń, Krystyna Pawłowicz, Katarzyna Lubnauer, Anna Maria Sierakowska, Łukasz Kohut, Marcin Kierwiński, Anna Maria Żukowska, Marian Banaś, Dariusz Joński, Kamila Gasiuk-Pihowicz, Barbara Nowacka, Adrian Zandberg, Krzysztof Śmieszek, Paulina Matysiak, Paweł Kukiz, Michał Wójcik, Sebastian Kaleta, Małgorzata Wassermann, Joachim Brudziński, Maciej Konieczny, Marcelina Zawisza

C) NGOs:

Polska Akcja Humanitarna, Helsińska Fundacja Praw Człowieka, Polski Czerwony Krzyż, Fundacja Dialog, Fundacja Ocalenie, Fundacja Ogólnopolski Strajk Kobiet, Stowarzyszenie Amnesty International, Fundacja Centrum Praw Kobiet, Stowarzyszenie Sędziów Polskich IUSTITIA, Stowarzyszenie Marsz Niepodległości, Lekarze bez Granic, Fundacja TVN, Fundacja Dzieciom "Zdążyć z Pomocą", Wielka Orkiestra Świątecznej Pomocy, Szlachetna Paczka, Fundacja WWF Polska, Fundacja Greenpeace Polska, Liga Ochrony Przyrody, Związek Stowarzyszeń Polska Zielona Sieć, Młodzieżowy Strajk Klimatyczny, Stowarzyszenie Miłość Nie Wyklucza, Kampania Przeciw Homofobii, Stowarzyszenie Lambda - Warszawa, Fundacja Trans-Fuzja, Stowarzyszenie Grupa Stonewall.

**3. Instructions for annotators**

You will evaluate the emotional content displayed in some short texts.

Your task will be to mark on a five-point scale the degree to which you think that a given sentence is characterized by each of the following emotions: joy, sadness, anger, disgust, fear, and pride. Use a five-point scale as described below:

0 - the emotion does not occur at all

1 - low level of emotion

2 - moderate level of emotion

3 - high level of emotion

4 - very high level of emotion.

Then, we will ask you to estimate the intensity of two additional emotion parameters: the direction of sensations (negative *versus* positive) and emotional arousal (no arousal *versus* extreme arousal). On the next screen you will learn the definitions of both parameters and how you will evaluate them.

Read the descriptions of two emotion parameters: the sign of sensations and emotional arousal. You can do this several times to make sure you understand them - it will make it easier for you to complete the task ahead of you.

You will rate each of the emotion dimensions described above on a five-point scale. To make it easier to imagine the states we have in mind, you can use pictograms symbolizing different directions of experiences and the intensity of the emotional states.

For the direction of sensations, use the following scale:

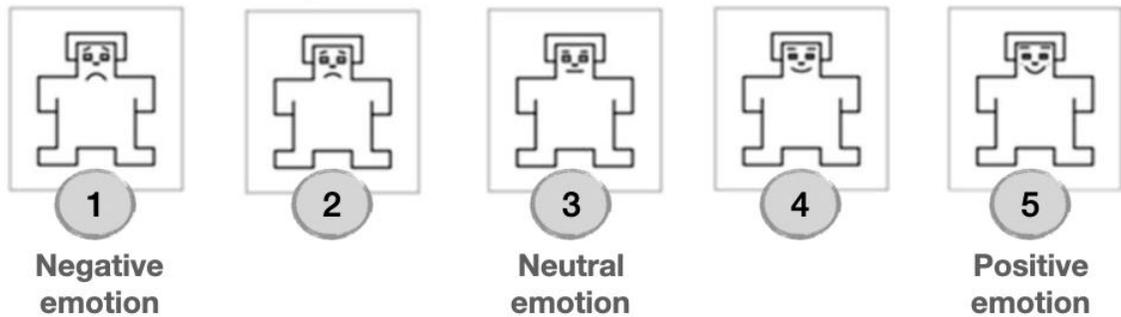

The first pictogram shows a person who is visibly depressed - specific experiences may include: panic, irritation, disgust, despair, failure, or crisis. The last image shows a person who is visibly excited - specific experiences may include: fun, delight, happiness, relaxation, satisfaction, or rest. The remaining pictograms represent intermediate states.

For emotional arousal, use the following scale:

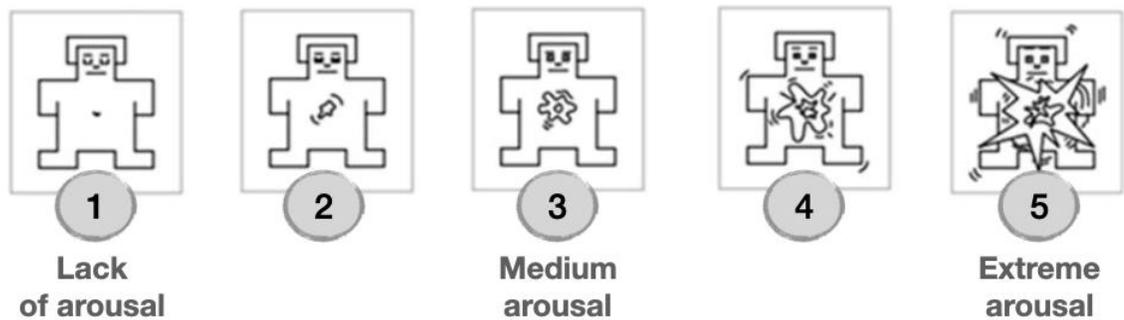

The first pictogram shows a person who is very calm, almost sleepy - specific experiences may include: relaxation, calm, inactivity, meditation, boredom, or laziness. The last image shows a person who is intensely aroused - appropriate emotional states may include: excitement, euphoria, arousal, rage, agitation, or anger.

Save the link to this manual for later - you can return to it at any time during the examination.

Very important: you can take a break while assessing your statements and return to them at any time - your current work will be saved and you will be able to resume it after the break. If you want to do this, in the upper right corner of the screen you will find the option: "Postpone for later" - click on it, enter the data necessary to save, and confirm the operation. In case you are ready to get back to work: when you enter the study page, an option "Load unfinished survey" will appear in the upper right corner of the screen - select it to load your work.

### 4. Politicians' names, Neutral and Political Sentences

Politicians' names:

Władysław Kosiniak-Kamysz, Szymon Hołownia, Andrzej Duda, Rafał Trzaskowski, Włodzimierz Czarzasty, Tomasz Grodzki, Mariusz Błaszczak, Elżbieta Witek, Donald Tusk, Mateusz Morawiecki, Jarosław Kaczyński, Zbigniew Ziobro, Sławomir Mentzen, Przemysław Czarnek, Robert Biedroń, Piotr Zgorzelski, Beata Szydło, Adrian Zandberg, Krzysztof Bosak, Michał Kołodziejczak

Table 1.

*Neutral Sentences*

| Original Sentence | Translation |
| --- | --- |
| [Name] poszedł do sklepu, aby kupić produkty spożywcze. | [Name] went to the store to buy groceries. |
| Przyszedł czas, aby [Name] wybrał film na wieczór. | It was time for [Name] to choose a movie for the evening. |
| [Name] zdecydował, że spotkanie zacznie się o | [Name] decided that the meeting would start at |
| dziesiątej. | ten o'clock. |
| Ulubionym kolorem [Name] jest niebieski. | [Name]'s favorite color is blue. |
| [Name] lubi czytać książki przed snem. | [Name] likes to read books before bedtime. |
| [Name] zjadł na lunch kanapkę i wypił filiżankę kawy. | [Name] had a sandwich and a cup of coffee for lunch. |
| [Name] zawsze jeździ do pracy autobusem. | [Name] always takes the bus to work. |
| W dzień wolny [Name] lubi odwiedzać lokalne muzeum. | On a day off, [Name] likes to visit the local museum. |

Table 2.

*Political Sentences*

| Original Sentence | Translation |
|---|---|
| [Name] opowiedział się za bardziej rygorystycznymi przepisami dotyczącymi ochrony środowiska podczas swojej kadencji. | [Name] advocated for stricter environmental protection regulations during their term. |
| [Name] sprzeciwił się nowej ustawie o reformie podatkowej, wyrażając obawy dotyczące jej wpływu na rodziny ze średnich warstw społecznych. | [Name] opposed the new tax reform bill, expressing concerns about its impact on middle-class families. |
| Podczas debaty [Name] obiecał zwiększyć finansowanie publicznej edukacji. | During the debate, [Name] promised to increase funding for public education. |
| [Name] jest zagorzałym zwolennikiem umów o wolnym handlu. | [Name] is a staunch supporter of free trade agreements. |
| [Name] skrytykował politykę zagraniczną rządu w ostatnim przemówieniu. | [Name] criticized the government's foreign policy in the latest speech. |
| Propozycja reformy służby zdrowia przedstawiona przez [Name] spotkała się z mieszanymi reakcjami różnych interesariuszy. | The healthcare reform proposal presented by [Name] received mixed reactions from various stakeholders. |
| [Name] konsekwentnie domaga się reform wyborczych mających na celu zwiększenie uczestnictwa wyborczego. | [Name] consistently calls for electoral reforms aimed at increasing voter participation. |
| W wywiadzie [Name] wyraził sceptycyzm co do skuteczności obecnych środków cyberbezpieczeństwa. | In an interview, [Name] expressed skepticism about the effectiveness of current cybersecurity measures. |

**5. Statistical analysis**

Table 1.

*Model with confounds only*

| | Dependent Variable: Valence of names | | |
|---|---|---|---|
| | (1) | (2) | (3) |
| Intercept | 47.383*** | 47.383*** | 47.383*** |
| | (0.770) | (0.724) | (0.941) |
| Gender | 7.789 | | |
| | (4.973) | | |
| Trust | | 2.155** | |
| | | (0.978) | |
| Mentions | | | 0.190* |
| | | | (0.092) |
| Observations | 22 | 22 | 22 |
| $R^2$ | 0.109 | 0.195 | 0.175 |
| Adjusted $R^2$ | 0.065 | 0.155 | 0.134 |
| Residual Std. Error | 3.553 (df=20) | 3.615 (df=20) | 3.200 (df=20) |
| F Statistic | 2.453 (df=1; 20) | 4.852** (df=1; 20) | 4.254** (df=1; 20) |

*Note*: *p<0.1; **p<0.05; ***p<0.01

## 6. Residuals' QQ-plots

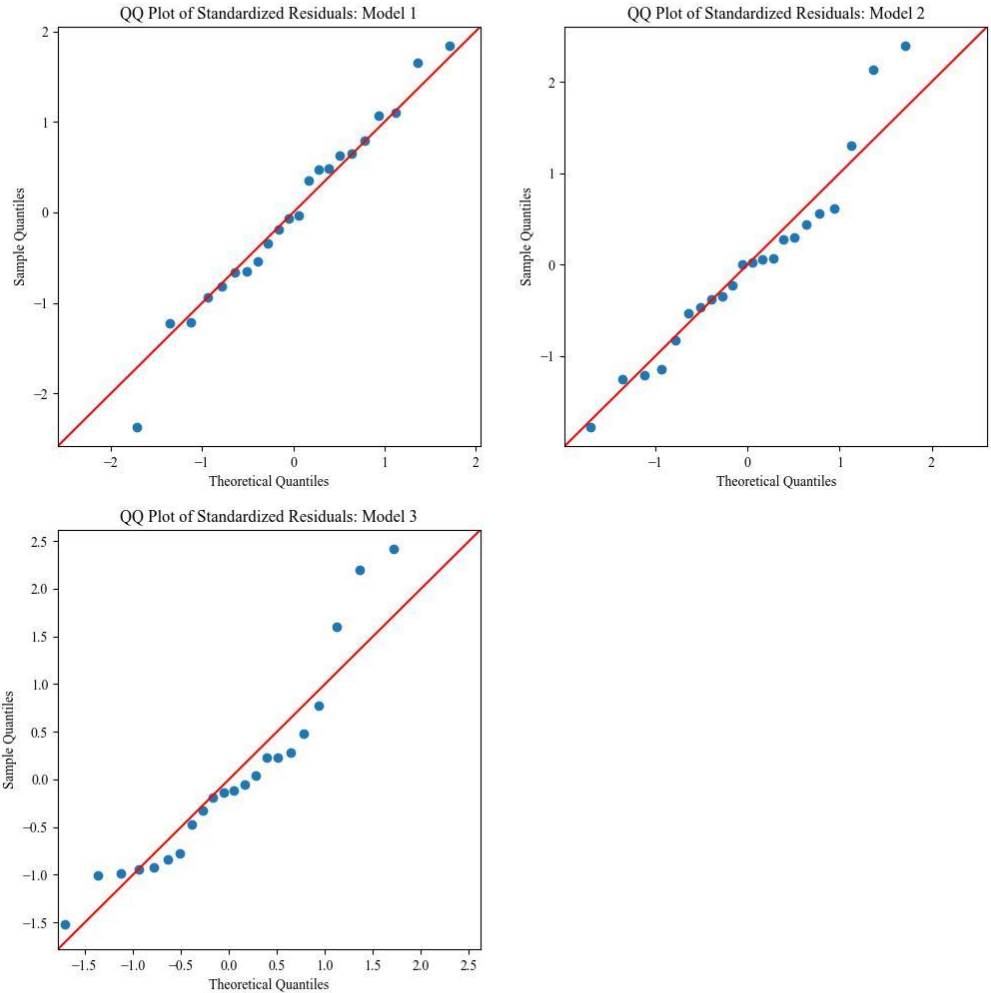

QQ plots of residuals for models in Table 1

QQ plots of residuals for models in Table 2

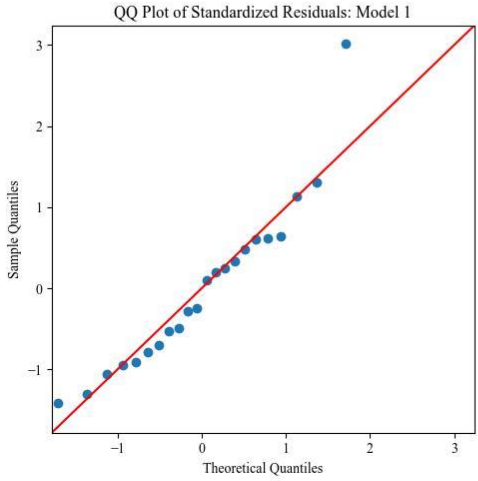
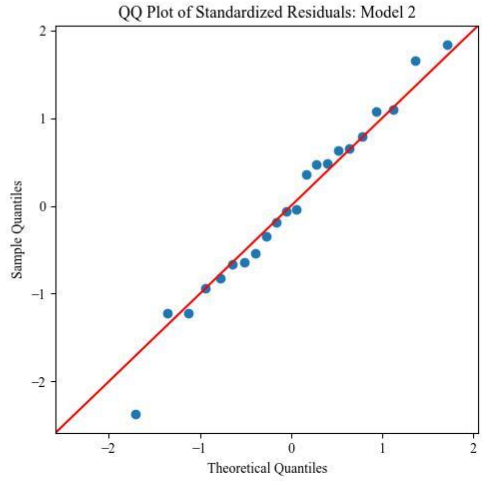
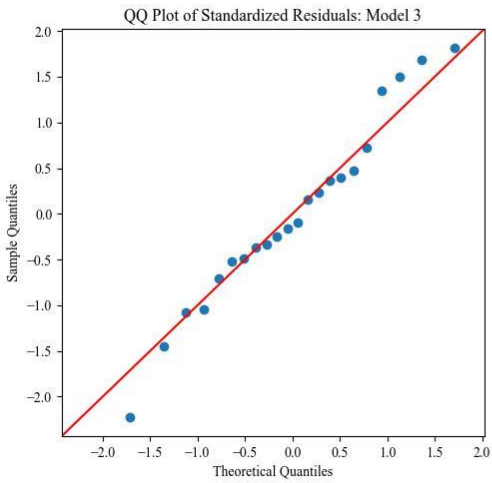
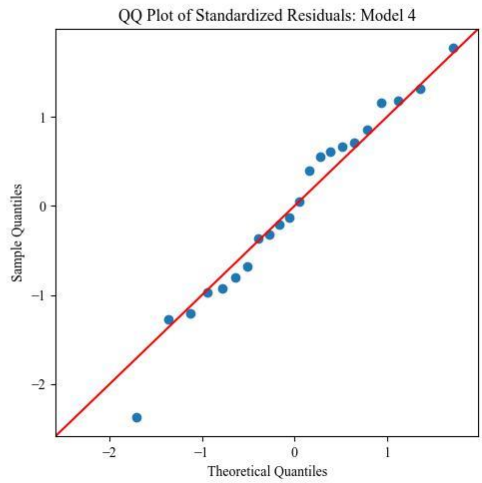
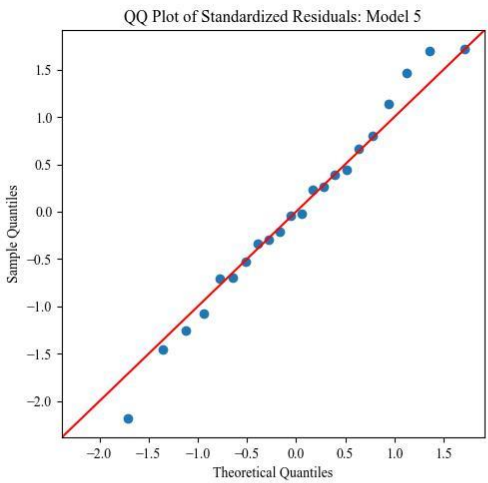

QQ plots of residuals for models in Table 3

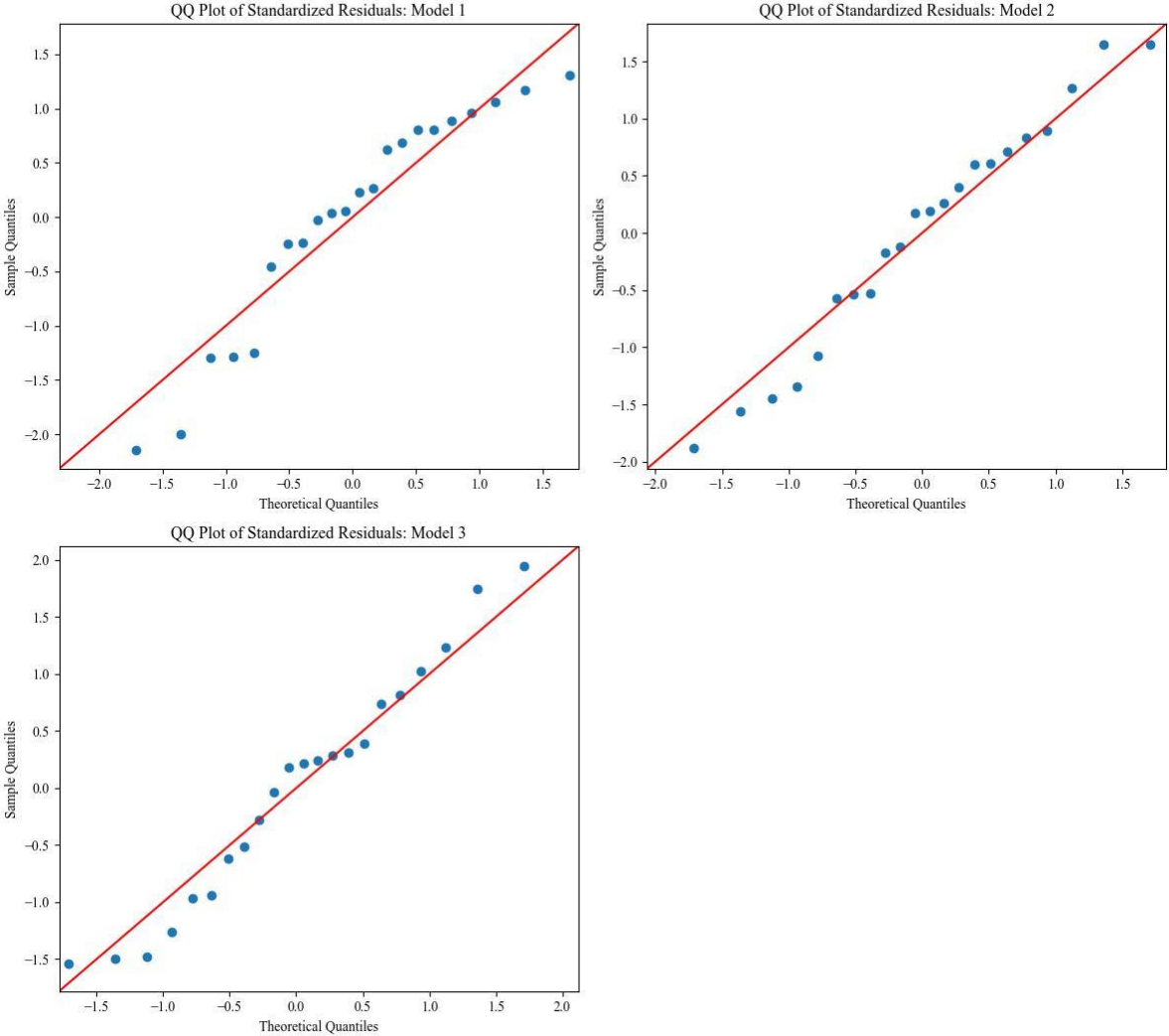